\pdfoutput=1

\documentclass[11pt]{article}

\usepackage[]{naacl2021}

\usepackage{times}
\usepackage{latexsym}

\usepackage[T1]{fontenc}

\usepackage[utf8]{inputenc}
\usepackage{microtype}
\usepackage{booktabs}
\usepackage{multirow}
\usepackage{graphicx}
\usepackage{caption}
\usepackage[normalem]{ulem}
\usepackage[utf8]{inputenc}
\usepackage{dingbat}
\usepackage{wrapfig}
\usepackage{amssymb}
\usepackage{amsfonts}
\usepackage{amsmath}
\usepackage{amssymb}
\usepackage{amsthm}
\usepackage{mathrsfs}
\usepackage{adjustbox}
\usepackage{bbold}
\usepackage[font=small]{caption}
\usepackage[linesnumbered,ruled]{algorithm2e}
\usepackage{url}
\usepackage{subcaption}
\usepackage{xspace}

\DeclareMathOperator*{\argmax}{argmax}

\renewcommand{\c}{\mathbf{c}}

\usepackage{enumitem}
\usepackage{wrapfig,lipsum,booktabs}

\usepackage[linesnumbered,ruled]{algorithm2e}
\usepackage{adjustbox}
\usepackage{mathtools}

\newcommand{\mvregfull}{Multi-view Subword Regularization\xspace}
\newcommand{\mvreg}{MVR\xspace}
\newcommand{\ps}{SR}

\usepackage[russian,greek,english]{babel}
\title{Multi-view Subword Regularization}

\author{
\textbf{Xinyi Wang}$^{1}$ \quad

\textbf{Sebastian Ruder}$^{2}$ \quad

\textbf{Graham Neubig}$^{1}$
\\
$^{1}$Language Technology Institute, Carnegie Mellon University \\
$^{2}$DeepMind \\
\texttt{xinyiw1@cs.cmu.edu,ruder@google.com,gneubig@cs.cmu.edu}
}

\begin{document}
\maketitle
\begin{abstract}
    Multilingual pretrained representations 
    generally rely on subword segmentation algorithms to create a shared multilingual vocabulary. However, standard heuristic algorithms often lead to sub-optimal segmentation, especially for languages with limited amounts of data. In this paper, we take two major steps towards alleviating this problem. First, we demonstrate empirically that applying existing subword regularization methods~\cite{subword_reg_kudo,bpe-dropout} during fine-tuning of pre-trained multilingual representations improves the effectiveness of cross-lingual transfer. Second, to take full advantage of different possible input segmentations, we propose \mvregfull~(\mvreg), a method that enforces the consistency between predictions of using inputs tokenized by the standard and probabilistic segmentations. Results on the XTREME multilingual benchmark~\cite{xtreme2020} show that \mvreg~brings consistent improvements of up to 2.5 points over using standard segmentation algorithms.\footnote{Code for the method is released here: \url{https://github.com/cindyxinyiwang/multiview-subword-regularization}}
    
\end{abstract}

\section{Introduction \label{sec:intro}}


Multilingual pre-trained representations  \citep{bert,uniencoder,xlm,xlmr} are now an essential component of state-of-the-art methods for cross-lingual transfer~\cite{wu-dredze-2019-beto,pires-etal-2019-multilingual}. These methods pretrain an encoder by learning in an unsupervised way
from raw textual data 
in up to hundreds of languages which can then be fine-tuned on annotated data of a downstream task in a high-resource language, often English, and transferred to another language.
In order to encode hundreds of languages with diverse vocabulary, it is standard for such multilingual models to employ a shared subword vocabulary jointly learned on the multilingual data using heuristic word segmentation methods based on byte-pair-encoding~\cite[BPE;][]{bpe} or unigram language models~\cite{sentencepiece} (details in \S\ref{sec:subword_reg}). However, subword-based preprocessing can lead to sub-optimal segmentation that is inconsistent across languages, harming cross-lingual transfer performance, particularly on under-represented languages. As one example, consider the segmentation of the word ``excitement'' in different languages in \autoref{tab:word_seg}. The English word is not segmented, but its translations in the other languages, including the relatively high-resourced French and German, are segmented into multiple subwords. Since each subword is mapped to a unique embedding vector, the segmentation discrepancy---which generally does not agree with a language's morphology---could map words from different languages to very distant representations, hurting cross-lingual transfer. 
In fact, previous work \cite{xlmr,artetxe2020cross} has shown that heuristic fixes such as increasing the subword vocabulary capacity and up-sampling low-resource languages during learning of the subword segmentation can lead to significant performance improvements.

\begin{table}[t]
    \centering
    \begin{tabular}{ll|ll}
    \toprule
       \textbf{en} & excitement & \textbf{fr} & excita\textcolor{red}{/}tion \\
       \textbf{de} & Auf\textcolor{red}{/}re\textcolor{red}{/}gung & \textbf{pt} & excita\textcolor{red}{/}\c{c}ão \\ 
          \textbf{el} & \begin{otherlanguage*}{greek}εν\textcolor{red}{/}θ\textcolor{red}{/}ουσι\textcolor{red}{/}ασμός\end{otherlanguage*} & \textbf{ru} & \begin{otherlanguage*}{russian}волн\textcolor{red}{/}ение \end{otherlanguage*} \\
    \bottomrule
    \end{tabular}
    \caption{XLM-R segmentation of ``excitement'' in different languages. The English word is not segmented while the same word in other languages is over-segmented. A better segmentation would allow the model to match the verb stem and derivational affix across languages.} 
    \label{tab:word_seg}
\end{table}

Despite this, there is not much work studying or improving subword segmentation methods for cross-lingual transfer. \citet{bpe_suboptimal} empirically compare several popular word segmentation algorithms for pretrained language models of a single language. Several works propose to use different representation granularities, such as phrase-level segmentation~\citep{ambert} or character-aware representations~\cite{charbert} for pretrained language models of a single high-resource language, such as English or Chinese only. However, it is not a foregone conclusion that methods designed and tested on monolingual models will be immediately applicable to multilingual representations. Furthermore, they add significant computation cost to the pretraining stage, which is especially problematic for multilingual pretraining on hundreds of languages. 
The problem of sub-optimal subword segmentation has drawn more attention in the context of neural machine translation~(NMT). Specifically, \emph{subword regularization} methods have been proposed to improve the NMT model of a \textit{single} language pair by randomly sampling different segmentations of the sentences during training \citep{subword_reg_kudo,bpe-dropout}. However, these methods have not been applied to multilingual NMT or pretrained language models and it is similarly not clear if they are useful for cross-lingual transfer. 

In this paper, we make two contributions to close this gap.
First, we perform the first (to our knowledge) empirical examination of subword regularization methods on a variety of cross-lingual transfer tasks from the XTREME benchmark~\cite{xtreme2020}.
We demonstrate that despite its simplicity, this method is highly effective, providing consistent improvements across a wide variety of languages and tasks for both multilingual BERT~\cite[mBERT;][]{bert} and XLM-R~\citep{xlmr} models.
Analysis of the results shows that this method is particularly effective for languages with non-Latin scripts despite only being applied during English fine-tuning. 

\begin{figure}
    \centering
    \includegraphics[width=\columnwidth]{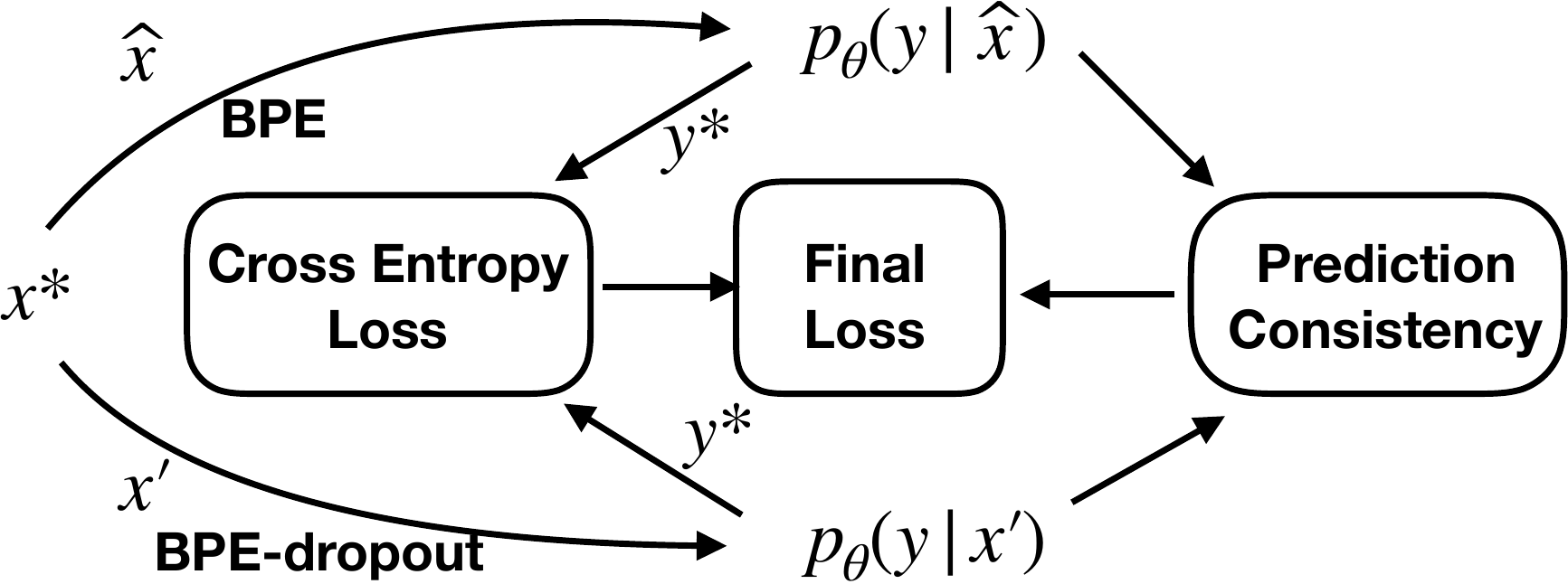}
    \caption{Fine-tuning models using \mvreg~on data $(x^*, y^*)$}
    \label{fig:mvr}
\end{figure}

Further, we posit that naively applying probabilistic segmentation only during fine-tuning may be sub-optimal as it creates a discrepancy between the segmentations during the pretraining and fine-tuning stages.
To address this problem, we propose \mvregfull~(\mvreg; \autoref{fig:mvr}), a novel method---inspired by the usage of consistency regularization in semi-supervised learning methods \cite{cvt,uda}---which utilizes \emph{both} the standard and probabilistically segmented inputs, enforcing the model's predictions to be consistent across the two views.
Such consistency regularization further improves accuracy, with \mvreg~finally demonstrating consistent gains of up to 2.5 points over the standard practice across all models and tasks. We analyze the sources of the improvement from consistency regularization and find that it can be attributed to both label smoothing and self-ensembling.


\section{Background: Subword Segmentation  \label{sec:subword_reg}}

Here, we first discuss two common deterministic segmentation methods based on byte pair encoding (BPE) and unigram language models (ULM), discuss their probabilistic variants, and explain how to incorporate them in training.

\subsection{Deterministic Segmentation}
The most widely used subword segmentation methods first estimate a segmentation model from the training corpus in an unsupervised fashion. They then produce a segmentation $\widehat{x}$ of the input $x^*$ under the estimated segmentation model $P(x)$:
\begin{align*}
\small
    \widehat{x} = \argmax_{x \in S(x^*)} P(x)
\end{align*}
Here $S(x^*)$ is the set of all possible segmentations, and $P(x)$ is the likelihood of a given segmentation. Note that $\widehat{x}$ is deterministically selected for each input $x^*$. 

\paragraph{Byte-pair encoding~(BPE)}
The popular BPE algorithm~\cite{bpe} initializes the vocabulary with individual characters and initially represents each word as a sequence of characters. It then counts the most frequent character token bigrams in the data, merges them into a new token, and adds the new token to the vocabulary. This process is done iteratively until a predefined vocabulary size is reached. 

To segment a word, BPE simply splits the word into character tokens, and iteratively merges adjacent tokens with the highest priority until no merge operation is possible. That is, for an input $x^*$, it assigns segmentation probability $P(\widehat{x})=1$ for the sequence $\widehat{x}$ obtained from the greedy merge operations, and assigns other possible segmentations a probability of 0.

Notably, a variant of this method~\cite{wordpiece} is used for the mBERT embedding model \citep{bert}.

\paragraph{Unigram language model~(ULM)} The ULM method~\cite{sentencepiece} starts from a reasonably large seed vocabulary, which is iteratively pruned to maximize the training corpus likelihood under a unigram language model of the subwords until the desired vocabulary size is reached. 

During segmentation, ULM decodes the most likely segmentation of a sentence under the estimated language model using the Viterbi algorithm. This method is used in the XLM-R cross-lingual embeddings \citep{xlmr}.

\subsection{Probabilistic Segmentation}
As explained in \S\ref{sec:intro}, one drawback of both word segmentation algorithms is that they produce a deterministic segmentation for each sentence, even though multiple segmentations are possible given the same vocabulary.
In contrast, \citet{subword_reg_kudo} and \citet{bpe-dropout} have proposed methods that enable the model to generate segmentations probabilistically. Instead of selecting the best subword sequence for input $x^*$, these method stochastically sample a segmentation $x'$ as follows:
\begin{align*}
\small
    x' \sim P'(x) \: \text{where} \: P'(x) \propto \begin{cases}
    P(x) \text{ if } x \in S(x^*) \\
    0 \text{ otherwise}
    \end{cases}
\end{align*}
Here we briefly introduce these two methods.
\paragraph{BPE-dropout} This method is used together with the BPE algorithm, randomly dropping merge operations with a given probability $p$ while segmenting the input data~\citep{bpe-dropout}. 
\paragraph{ULM-sample} As the ULM algorithm relies on a language model to score segmentation candidates for picking the most likely segmentation, \citet{subword_reg_kudo} propose to sample from these segmentation candidates based on their language model scores.

\subsection{Subword Regularization~(SR)}

Subword regularization \citep{subword_reg_kudo} is a method that incorporates probabilistic segmentation at training time to improve the robustness of models to different segmentations.
The idea is conceptually simple: at training time sample different segmentations $x'$ for each input sentence $x^*$.
Previous works \citep{subword_reg_kudo,bpe-dropout} have demonstrated that subword regularization using both BPE-dropout and ULM-sampling are effective at improving machine translation accuracy, particularly in cross-domain transfer settings where the model is tested on a different domain than the one on which it is trained.

\section{Subword Regularization for Cross-lingual Transfer}

\begin{figure}
    \centering
    \includegraphics[width=0.7\columnwidth]{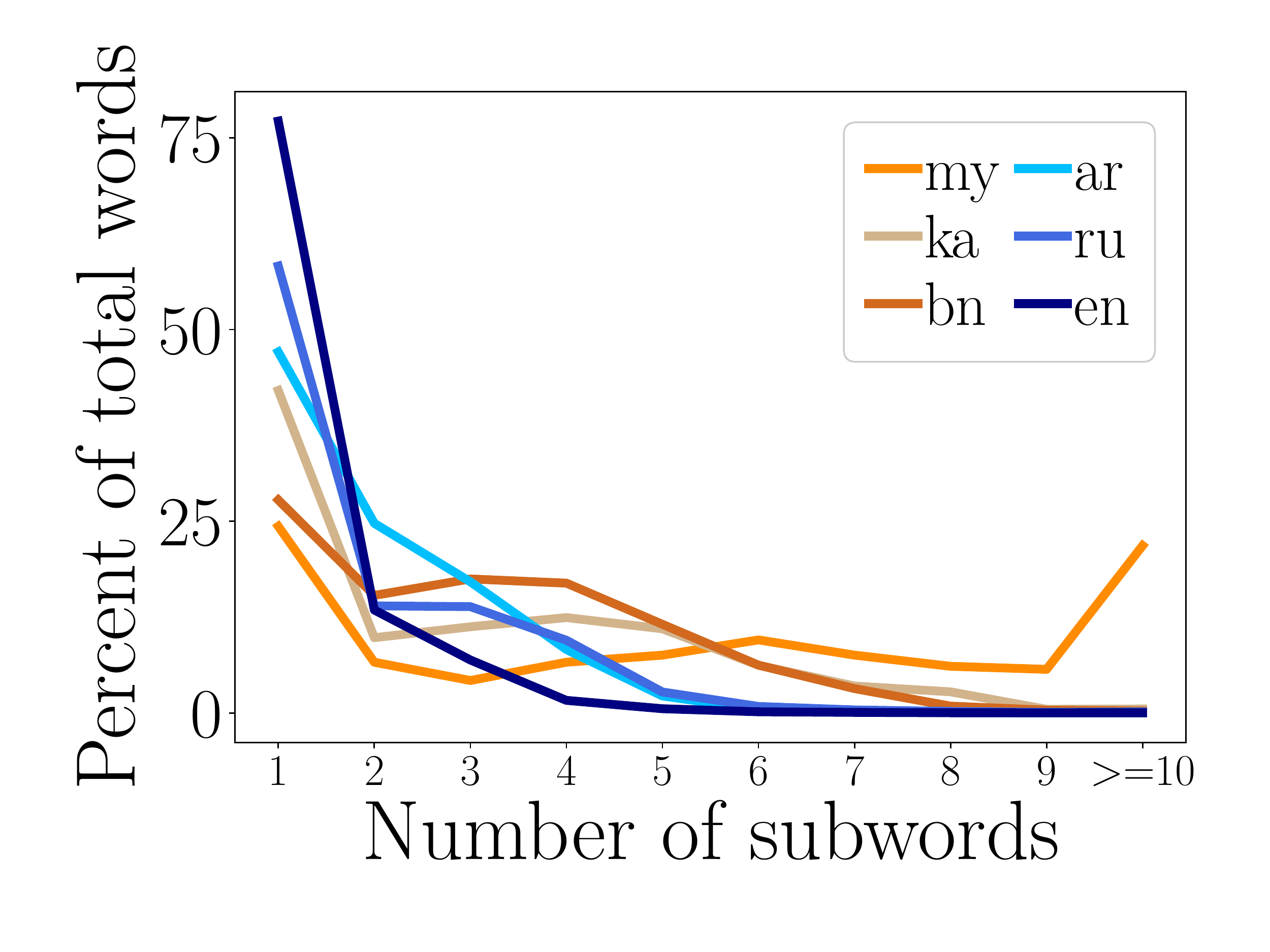}
    \caption{Percentage of words with different number of segments from different languages.}
    \label{fig:word_segment}
\end{figure}

While sub-optimal word segmentation is a challenge in monolingual models, it is an even bigger challenge for multilingual pretrained models.
These models train a shared subword segmentation model jointly on data from many languages, but the segmentation can nonetheless be different across languages, stemming from two main issues.
First, the granularity of segmentation differs among languages, where the segmentation model tends to \textit{over-segment} low-resource languages that do not have enough representation in the joint training data~\citep{acs_2019}.    
\autoref{fig:word_segment} shows the distribution of words from languages from different language families based on the number of subwords they are split into.\footnote{We use \citet{Pan2017}'s named entity recognition test data with mBERT's tokenizer.} We can see that the majority of English words are not segmented at all, while many languages only have less than half of the words unsegmented. Notably, even though Burmese~(my) is a language with little inflectional morphology, almost a quarter of the words are segmented into more than nine subwords.
Second, the segmentation might still be \textit{inconsistent} between different languages even if the granularity is similar, as explained in \autoref{tab:word_seg}. For example, neither the English word ``excitement'' nor the same word in French ``excita/tion'' are overly segmented, but segmenting the English word into ``excite/ment'' would allow the model to learn a better cross-lingual alignment.

Despite these issues, few methods have tried to address this subword segmentation problem for multilingual pretrained models. \citet{chau-etal-2020-parsing} propose to adapt a pretrained multilingual model to a new language by augmenting the vocabulary with a new subword vocabulary learned on the target language, but this method might not help for languages other than the target language it adapts to. \citet{cluster_vocab_multilingual} propose to separately construct a subword segmentation model for each cluster of related languages for \textit{pretraining} the multilingual representations. However, directly modifying the word segmentation requires retraining large pretrained models, which is computationally prohibitive in most cases.

In this paper, we instead propose a more efficient approach of using probabilistic segmentation during \textit{fine-tuning} on labeled data of a downstream task.
As mismatch in segmentation is one of the factors harming cross-lingual transfer, we expect a model that becomes more robust to different varieties of segmentation in one language will be more accommodating to differing segmentations in other languages during inference. 
Despite the simplicity of this method it is, as far as we are aware, unattested in the literature, and we verify in \autoref{sec:main_result} that it significantly improves the cross-lingual transfer performance of multilingual pretrained models.



\section{\mvregfull\label{sec:kl_reg}}

Previous attempts at SR have mainly applied it to models trained \emph{from scratch} for tasks such as MT.
However, the situation is somewhat different when fine-tuning pre-trained representations, in which case the original pre-trained models are generally not trained on sampled segmentations.
This discrepancy between the segmentation of the English labeled data and the segmentation of English monolingual data during pretraining might hurt the ability of the model to take full advantage of the parameters learned during the pretraining stage. 
To reduce this pretraining--fine-tuning discrepancy,
we propose \mvregfull~(\mvreg), a method for learning from multiple segmented versions of the same data and enforcing the consistency of predictions over different segmentations. 

Given the input $\widehat{x}_i$ tokenized with the deterministic segmentation such as BPE, and $x'_i$, the same input tokenized with the corresponding probabilistic segmentation algorithm such as BPE-dropout, the objective for \mvreg~has three components
\begin{multline}
\label{eqn:obj_full}
    J(\theta) = \sum_{i=1}^n \big[ -\frac{1}{2} \underbrace{\text{log} \: p_\theta(y_i|\widehat{x}_i)}_\text{Det. Seg CrossEnt} - \frac{1}{2} \underbrace{\text{log} \: p_\theta(y_i|x'_i)}_\text{Prob. Seg CrossEnt} \\
     + \lambda \underbrace{D(p_\theta(y_i|\widehat{x}_i) \: || \: p_\theta(y_i|x'_i))}_\text{Consistency loss} \big]
\end{multline}
\begin{enumerate}[leftmargin=*]
\itemsep0em 
    \item A cross-entropy loss using the standard deterministic segmentation. This loss acts on data whose segmentation is consistent with the segmentation seen during pretraining. It thus maximizes the benefit of pretrained representations.
    \item A cross entropy loss using probabilistic segmentation. It allows the model to learn from different possible segmentations of the same input.  
    \item A distance term $D(\cdot \:|| \: \cdot)$ between the model prediction distributions over the two different versions of the input. We use KL divergence as the distance metric and a hyperparameter $\lambda$ to balance the supervised cross-entropy losses and the consistency loss. Minimizing the distance between the two distributions enforces the model to make consistent predictions under different input segmentations, making it robust to sub-optimal segmentation of multilingual data.\footnote{As in semi-supervised learning \cite{cvt}, we expect our method to also be effective when applied to unlabeled data, e.g. using target language adaptation \cite{Pfeiffer2020mad_x}, which we leave for future work.}
\end{enumerate}

\paragraph{Flattening the prediction} The benefit of consistency regularization might be limited if the model prediction becomes overly confident on certain classes, especially when the number of output classes is large. Inspired by a similar technique in knowledge distillation~\citep{knowledge_distill}, we can use a softmax temperature $\tau$ to flatten the prediction distribution when calculating the consistency loss. Specifically, the distance loss between two prediction distributions in \autoref{eqn:obj_full} can be written as $D(p^{\text{flat}}_\theta(y_i|\widehat{x}_i) \: || \: p_\theta(y_i|x'_i))$, where
\begin{align}
    p^{\text{flat}}_\theta(y_i|\widehat{x}_i) = \frac{\text{exp}(z_y)/\tau}{\sum_{y'} \text{exp}(z_{y'})/\tau}
\end{align}
and $z_y$ is the logit for output label $y_i$. Normally $\tau$ is set to 1, and a higher $\tau$ makes the probability distribution more evenly distributed over all classes. In our experiments, we find that $\tau=1$ works well for most of the tasks and $\tau=2$ works slightly better for tasks that have larger output label spaces.

\paragraph{Efficiency} At inference time, we simply use the model prediction based on the input tokenized by deterministic segmentation only. Therefore, our method does not add additional decoding latency. \mvreg~needs about twice the fine-tuning cost compared to the baseline. However, compared to pretraining and inference usage of a model, fine-tuning is generally the least expensive component.

\section{Experiments\label{sec:exp}}

\subsection{Training and evaluation}
We evaluate the multilingual representations using tasks from the XTREME benchmark~\cite{xtreme2020}, focusing on the zero-shot cross-lingual transfer with English as the source language. We consider sentence classification tasks including XNLI~\cite{Conneau2018xnli} and PAWS-X~\cite{Yang2019paws-x}, a structured prediction task of multilingual NER~\citep{Pan2017}, and question-answering tasks including XQuAD~\citep{artetxe2020cross} and MLQA~\citep{Lewis2020mlqa}.

\subsection{Experiment setup}
We evaluate on both the mBERT model which utilizes BPE to tokenize the inputs, and the XLM-R models which uses ULM segmentation. To replicate the baseline, we follow the hyperparameters provided in the XTREME codebase\footnote{\url{https://github.com/google-research/xtreme}}.
Models are fine-tuned on English training data and zero-shot transferred to other languages. We run each experiment with 5 random seeds and record the average results and the standard deviation. 

\paragraph{\ps}
We use BPE-dropout~\cite{bpe-dropout} for mBERT and ULM-sample~\cite{subword_reg_kudo} for XLM-R models to do probabilistic segmentation of the English labeled data. BPE-dropout sets a dropout probability of $p \in [0, 1]$ for the merge operations, where a higher $p$ corresponds to stronger regularization. ULM-sample  utilizes a sampling temperature $\alpha \in [0, 1]$ to scale the scores for segmentation candidates, and a lower $\alpha$ leads to stronger regularization. We select the $p$ and $\alpha$ values based on the model performance on the English dev set of the NER task and simply use the same values across all other tasks. We set $p = 0.1$ for BPE-dropout and $\alpha = 0.6$ for ULM-sample.    

\paragraph{\mvreg} 
We select the hyperparameters for \mvreg~using the English dev set performance on the NER task. \mvreg~works slightly better by using stronger regularization than \ps, likely because using inputs deterministically segmented by the standard algorithm can balance the negative impact of bad tokenization by sampling from a more diverse set of segmentation candidates. 
We use $\lambda=0.2, p=0.2$ for mBERT and $\lambda=0.6, \alpha=0.2$ for XLM-R. We use prediction temperature $\tau=2$ for the question-answering tasks XQuAD and MLQA for the XLMR mdoels, and simply use $\tau=1$ for all other tasks. Further analysis of hyperparameters on the performance of \mvreg~can be found in \autoref{app:hyperparam}.

\subsection{Main results\label{sec:main_result}}
\newcommand{\rpm}{\raisebox{.2ex}{$\scriptstyle\pm$}}
\begin{table*}[htbp]
    \centering
    \small
    \resizebox{\textwidth}{!}{%
    \begin{tabular}{ll|llllll}
    \toprule
      Model & Method & Avg. & XNLI & PAWS-X & NER & XQuAD & MLQA \\
      \midrule
      Metrics & & & Acc. & Acc. & F1 & F1/EM & F1/EM \\
      \midrule
      \multirow{3}{*}{mBERT} & \citet{xtreme2020} & 67.1 & 65.4 & 81.9 & 62.2 & 64.5 / 49.4 & 61.4 / 44.2 \\
      & Baseline~(ours) & 67.3 & 66.5\rpm{0.4} & 83.1\rpm{0.4} & 61.5\rpm{0.7} & 64.7\rpm{0.2} / 49.8\rpm{0.4} & 60.9\rpm{0.4} / 43.8\rpm{0.5} \\
       & \ps & 68.0 & 66.4\rpm{0.2} & 85.0\rpm{0.3} & 62.2\rpm{0.6} & 64.7\rpm{0.3} / 50.0\rpm{0.3} & 61.5\rpm{0.3} / 44.4\rpm{0.3} \\
       & \mvreg & \textbf{68.8} & \textbf{67.2}\rpm{0.3} & \textbf{85.6}\rpm{0.3} & \textbf{62.7}\rpm{0.4} & \textbf{66.3}\rpm{0.2} / \textbf{51.7}\rpm{0.2} & \textbf{62.2}\rpm{0.2} / \textbf{45.3}\rpm{0.1} \\
      \midrule
       \multirow{3}{*}{XLM-R base} & Baseline~(ours) & 71.1 & 74.4\rpm{0.2} & 84.3\rpm{0.7} & 60.6\rpm{0.6} & 70.9\rpm{0.3} / 54.9\rpm{0.5} & 65.5\rpm{0.3} / 47.7\rpm{0.2} \\
       & \ps  & 71.4 & 74.4\rpm{0.7} & 85.5\rpm{0.5} & 61.0\rpm{0.6} & 70.9\rpm{0.3} / 55.7\rpm{0.2} & 65.4\rpm{0.1} / 47.5\rpm{0.1} \\
       & \mvreg & \textbf{72.3} & \textbf{75.3}\rpm{0.3} & \textbf{86.3}\rpm{0.6} & \textbf{61.8}\rpm{0.3} & \textbf{71.6}\rpm{0.5} / \textbf{56.5}\rpm{0.4} & \textbf{66.4}\rpm{0.5} / \textbf{48.5}\rpm{0.4} \\
      \midrule
        \multirow{3}{*}{XLM-R large} & \citet{xtreme2020} & 75.8 & 79.2 & 86.4 & 65.4 & 76.6 / 60.8 & 71.6 / 53.2 \\
        & Baseline~(ours) & 76.1 & 80.3\rpm{0.4} & 86.9\rpm{0.5} & 63.6\rpm{0.3} & 77.0\rpm{0.2} / 61.7\rpm{0.3} & \textbf{72.8}\rpm{0.2} / \textbf{54.5}\rpm{0.1} \\
       & \ps & 76.5 & 80.1\rpm{0.5} & 87.3\rpm{0.4} & 65.5\rpm{0.6} & 77.2\rpm{0.2} / 62.0\rpm{0.2} & 72.5\rpm{0.1} / 54.0\rpm{0.2} \\
       & \mvreg & \textbf{77.2} & \textbf{81.3}\rpm{0.1} & \textbf{88.2}\rpm{0.2} & \textbf{66.0}\rpm{0.7} & \textbf{77.6}\rpm{0.2} / \textbf{62.5}\rpm{0.4} & \textbf{72.8}\rpm{0.2} / \textbf{54.5}\rpm{0.1} \\     
    \bottomrule
    \end{tabular}
    }
    \caption{Average performance and standard deviation of different methods for mBERT, XLM-R base and XLM-R large models. SR is especially effective for mBERT. \mvreg~leads to significant further improvements across all models and tasks.}
    \label{tab:main_result}
\end{table*}

We compare performance of \ps, \mvreg~and the baseline for all models in \autoref{tab:main_result}, focusing on the average performance on all languages for each task. Our baseline numbers match or exceed the benchmark results in \citet{xtreme2020} for both mBERT and XLM-R large~(\citet{xtreme2020} do not include results for XLM-R base) on almost all tasks. 

\paragraph{Applying \ps~on English significantly improves other languages}
\ps~is surprisingly effective for mBERT---it is comparable to the baseline on XNLI and significantly improves over the baseline for the rest of the four tasks. However, the gains are less consistent for XLM-R models.
For both XLM-R base and large, \ps~leads to improvements on the NER task and the PAWS-X classification task, but is mostly comparable to the baseline for the rest of the three tasks.
\ps~performs better for mBERT likely because the vocabulary of mBERT is more imbalanced than that of XLM-R; it thus benefits more from the regularization methods. mBERT relies on BPE, which could be worse than ULM at tokenizing subwords into morphologically meaningful units~\citep{bpe_suboptimal}. Furthermore, mBERT has only 100K words in the vocabulary while XLM-R has a much larger vocabulary of 250K.  

\paragraph{\mvreg~consistently improves over \ps}
For mBERT, it leads to improvements of over 1 to 2 points over the baseline for all tasks. It is also very effective for the XLM-R models. For both the XLM-R base and the stronger XLM-R large models, \mvreg~improves over 1 point over the baseline on the NER task and the two classification tasks. On the question-answering tasks, \mvreg~delivers strong improvements for the XLM-R base model while the improvements on the XLM-R large model is slightly smaller. It has around 0.5 point improvement on XQuAD and has the same performance on MLQA. \mvreg~leads to more improvements on XQuAD, probably because it has a more diverse set of languages that potentially have more sub-optimal subword segmentation. The consistent gains on both mBERT and XLM-R show that \mvreg~is a general and flexible method for a variety of pretrained multilingual models based on different segmentation methods.

\subsection{Effect of each loss component}
In this section, we verify the effectiveness of the three loss components in \mvreg~by removing each of them from the objective. The ablation results on mBERT for all tasks are listed in \autoref{tab:ablate}. Removing any of the three loss components hurts the model performance by about the same amount for most of the tasks. For the question answering tasks, however, removing the cross-entropy loss on the deterministically segmented inputs reduces the model performance by almost half. This is likely because under this setting, the model only learns to locate exact spans for inputs tokenized by BPE-dropout, while we use the standard BPE to segment the inputs at test time. 
\begin{table*}[]
    \centering
    \small
    \resizebox{0.9\textwidth}{!}{%
    \begin{tabular}{l|llllll}
    \toprule
       Method & Avg.  & XNLI & PAWS-X & NER & XQuAD & MLQA \\
      Metrics &   & Acc. & Acc. & F1 & F1/EM & F1/EM \\
      \midrule
       \mvreg & 68.8 & 67.2\rpm{0.3} & 85.6\rpm{0.3} & 62.7\rpm{0.4} & 66.3\rpm{0.2} / 51.7\rpm{0.2} & 62.2\rpm{0.2} / 45.3\rpm{0.1} \\
       \midrule
        $\:$ -- Det. Seg CrossEnt & 52.8 & 66.7\rpm{0.5} & 85.5\rpm{0.2} & 62.4\rpm{0.7} & 25.0\rpm{8.2} / 15.6\rpm{6.7}  & 24.3\rpm{8.6} / 13.1\rpm{6.4} \\
        $\:$ -- Prob. Seg~CrossEnt & 67.7 & 66.7\rpm{0.6} & 85.0\rpm{0.3} & 62.3\rpm{0.7} & 64.0\rpm{0.3} / 48.7\rpm{0.4} & 60.3\rpm{0.4} / 43.1\rpm{0.3}  \\
        $\:$ -- consistency loss & 68.2 & 66.5\rpm{0.7} & 85.3\rpm{0.3} & 62.3\rpm{0.6} & 65.2\rpm{0.2} / 50.2\rpm{0.4} & 61.7\rpm{0.1} / 44.5\rpm{0.2} \\
    \bottomrule
    \end{tabular}
    }
    \caption{Effect of removing each loss component on mBERT.}
    \label{tab:ablate}
\end{table*}

\section{Analyses \label{sec:ana}}

In this section, we perform several analyses to better understand the behavior and root causes of the accuracy gains realized by our method.


\begin{figure*}[!t]
    \centering
    \begin{minipage}[t]{0.305\textwidth}
        \centering
        \includegraphics[width=0.98\linewidth]{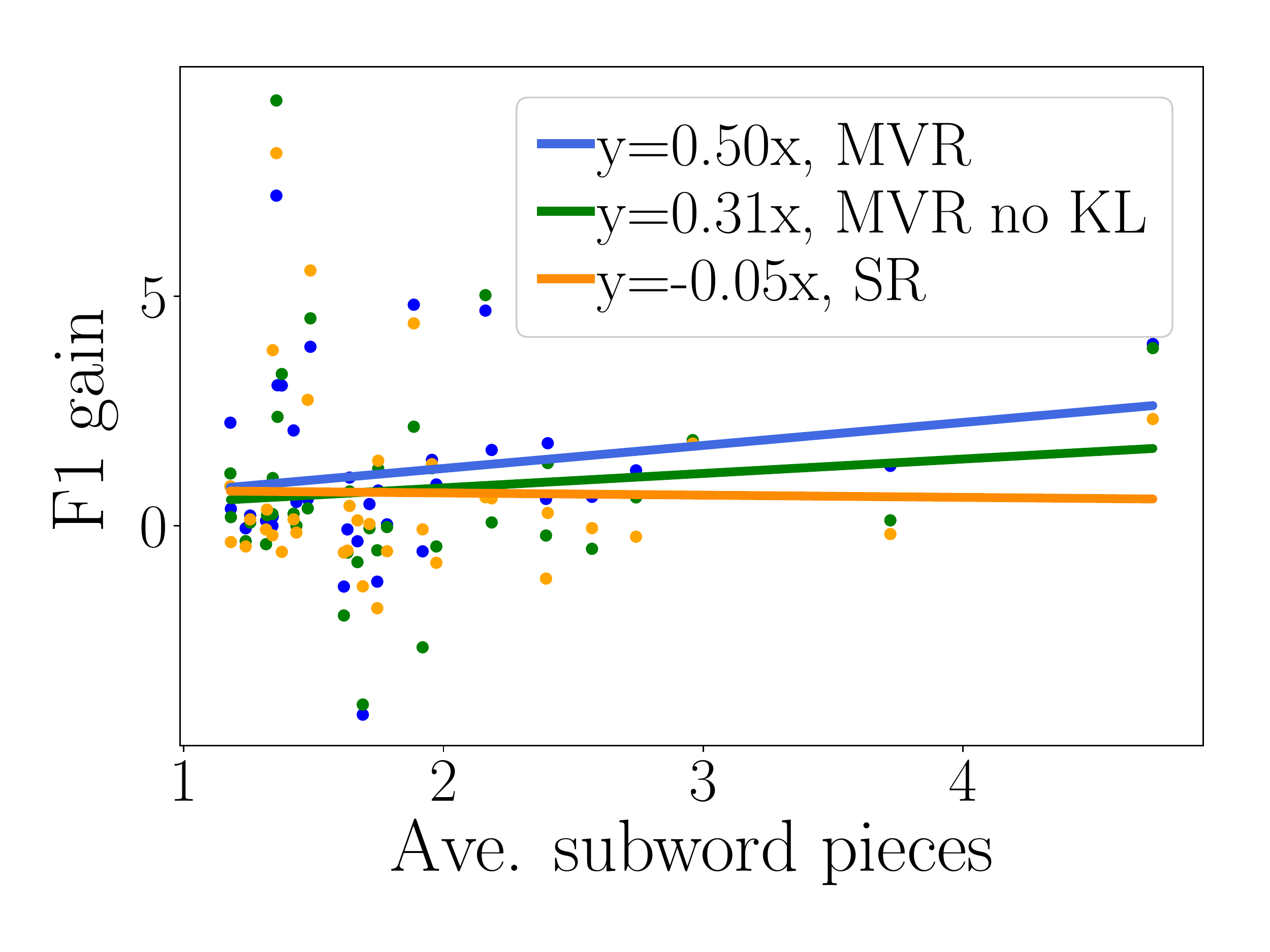}
        \vspace{-0.7em}
        \caption{mBERT gains over the NER baseline by average word pieces of a language. \mvreg~tends to benefit over-segmented languages more.}
        \label{fig:gain_vs_seglang}
    \end{minipage}
    \hspace{5mm}
    \begin{minipage}[t]{0.305\textwidth}
        \centering
        \includegraphics[width=0.98\linewidth]{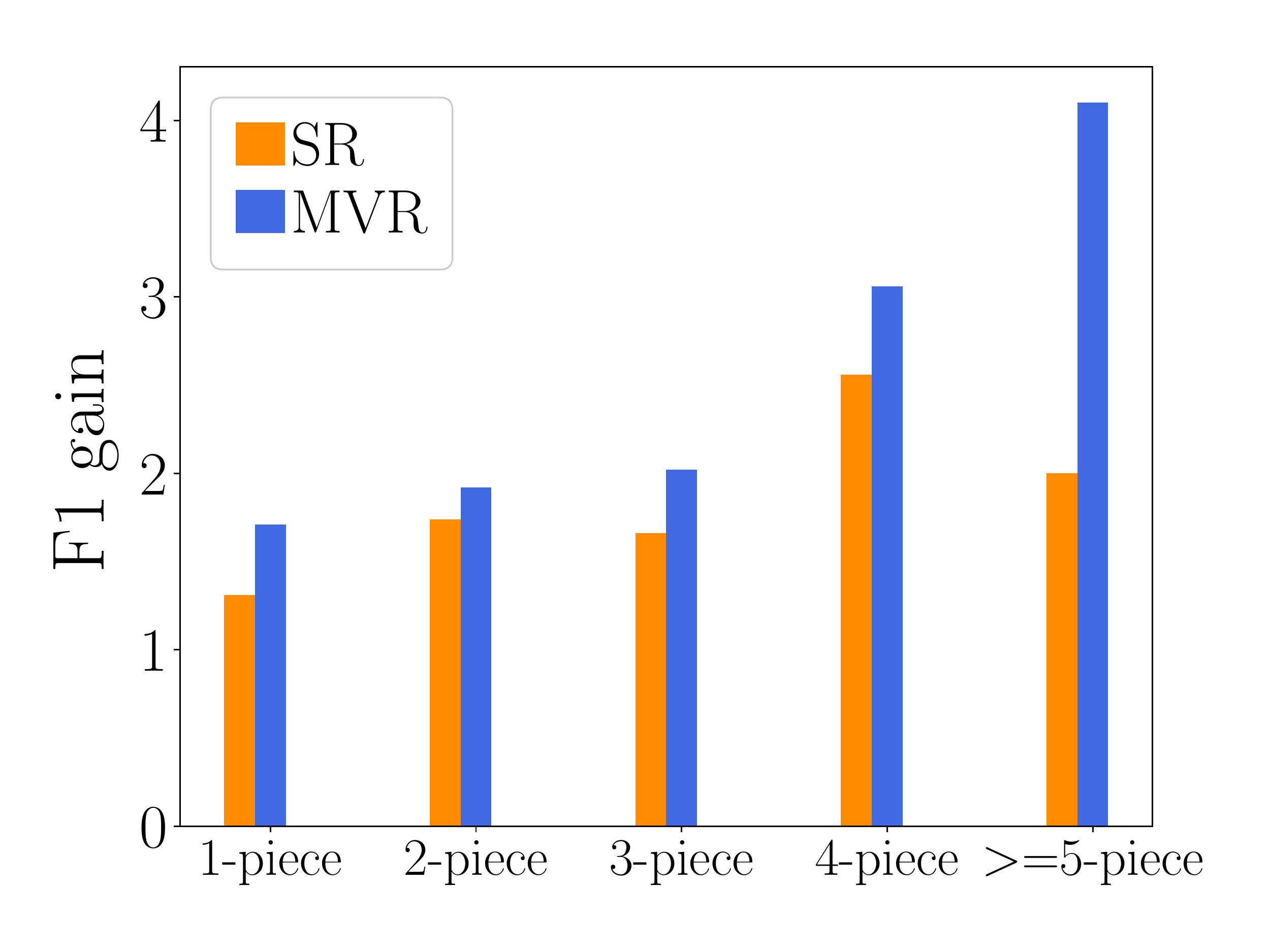}
        \vspace{-0.7em}
        \caption{XLM-R large gains over the NER baseline for words with increasing number of subword pieces.}
        \label{fig:gain_vs_segword}
    \end{minipage}
    \hspace{5mm}
    \begin{minipage}[t]{0.305\textwidth}
        \centering
        \includegraphics[width=0.98\linewidth]{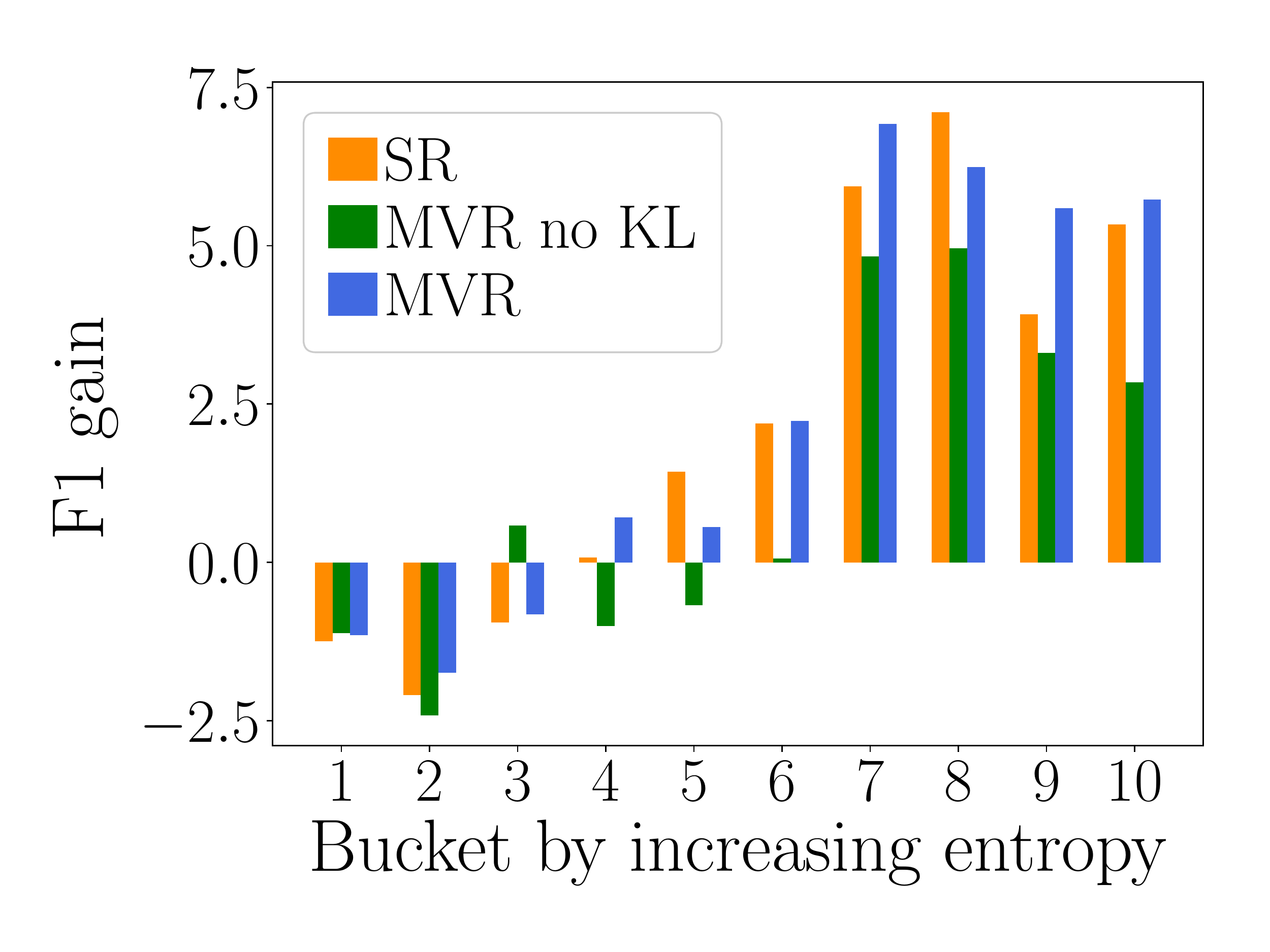}
        \vspace{-0.7em}
        \caption{mBERT improvements over the NER baseline for examples with different entropy. Consistency loss helps examples with higher entropy more.}
        \label{fig:gain_entropy}
    \end{minipage}
\end{figure*}

\begin{figure*}[!t]
    \centering
    \begin{minipage}[t]{0.6\textwidth}
        \centering
        \includegraphics[width=\columnwidth]{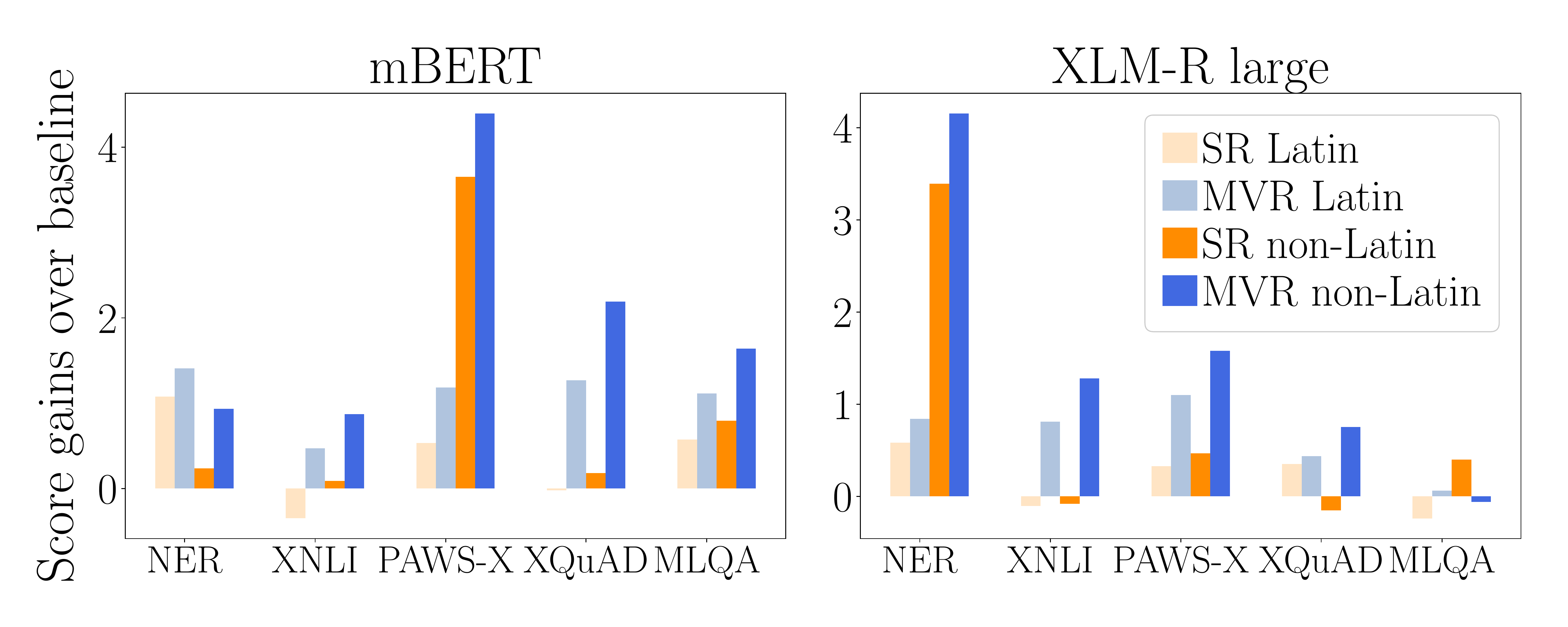}
        \vspace{-0.7em}
        \caption{Gains over the baseline for languages with Latin vs. non-Latin script. Both \ps~and \mvreg~improve more for non-Latin languages.} 
        \label{fig:gain_for_script}
    \end{minipage}
    \hspace{5mm}
    \begin{minipage}[t]{0.35\textwidth}
        \centering
        \includegraphics[width=0.98\linewidth]{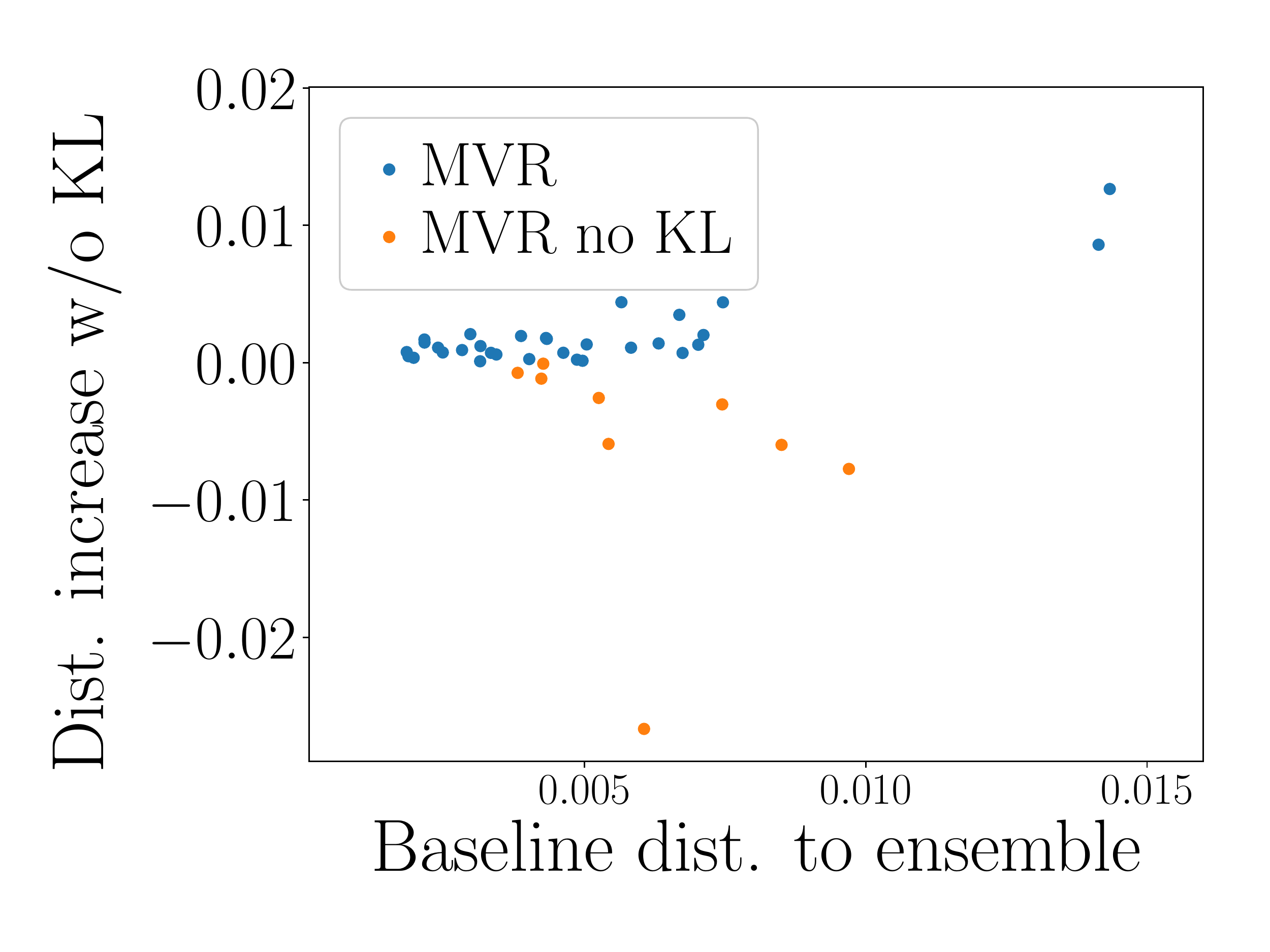}
        \vspace{-0.7em}
       \caption{Increase in distance to the ensemble distribution by removing the consistency loss on the NER task. The languages are labeled based on the method with closer distance to the ensemble distribution. The consistency loss shifts model prediction closer to the ensemble distribution.}
       \label{fig:ensemble}
    \end{minipage}
\end{figure*}

\subsection{Effect on over-segmentation}
In this section, we analyze the effect of our methods on languages and words with different subword segmentation granularity. We focus on the NER task because it contains a diverse set of over 40 languages. We calculate the average number of subword pieces in a language, and plot the gains over the baseline for these languages with respect to their average subwords in \autoref{fig:gain_vs_seglang}. To visualize the relationship between the two values, we also fit a trend line and record its coefficient for each method in the legend. We consider three methods for mBERT: \ps, \mvreg~without consistency loss, and the full \mvreg. The trend line for \mvreg~has a positive coefficient, indicating that it improves more on languages that are more overly segmented. Removing the consistency loss tends to hurt more for these languages. \ps, on the other hand, does not tend to favor languages with more subword segmentation.

Next, we bucket all the words together based on how many subwords they are segmented into, and compare the performance of our methods for each word bucket. We use the XLM-R model and plot the results in \autoref{fig:gain_vs_segword}. \ps~brings slightly more improvements on average for words that are split into 4 or more pieces for the large model. \mvreg~outperforms \ps~for all categories, especially for difficult words that are segmented into 5 or more subwords.

\paragraph{Gains on Latin vs. non-Latin script}
In addition, it is notable that we fine-tune the model using labeled data from English, a Latin script language, while the non-Latin scripted languages might have larger segmentation and vocabulary discrepancies from English.
We thus also plot the score improvements of both \ps~and \mvreg~over the baseline for languages with and without Latin script in \autoref{fig:gain_for_script}. We use a lighter shade to represent improvements for Latin-script languages and a darker shade for languages with non-Latin scripts. Across all the tasks, both \ps~and \mvreg~generally have larger improvements on languages with non-Latin script. \mvreg, which is represented by blue shades, generally outperforms \ps~for both the Latin and non-Latin scripted languages across all models. While \ps~sometimes underperforms the baseline on Latin scripted languages, especially for XLM-R models, \mvreg~delivers consistent improvements over the baseline across both types of languages. Overall, \mvreg achieves the largest improvements over \ps~for languages with non-Latin scripts.

\subsection{Effect of consistency loss}
One of the novel components of \mvreg~is the consistency loss between two different segmentations of the input. In this section we analyze two hypotheses about the source of benefit provided thereby. 
\paragraph{Label smoothing}
The first hypothesis is that the consistency loss may be able to mitigate over-confident predictions by calibrating the two output distributions against each other.
This effect is similar to label smoothing~\citep{label_smoothing,revisit_kd}, which softens the one-hot target label by adding a loss of uniform distribution over all class labels and has proven helpful across a wide variety of models.
To measure this, we plot the F1 improvement on the NER task for examples categorized by increasing predictive entropy in \autoref{fig:gain_entropy}.
\mvreg~leads to more improvements on examples with higher entropy, or those that the model is more uncertain about, indicating that MVR is indeed helping the model improve on examples where it is not confident.

\paragraph{Ensemble effect}
The second hypothesis is that the consistency loss could regularize the model to be closer to the ensemble of models trained on standard deterministically segmented inputs and probabilistically segmented inputs. To verify this hypothesis, we first calculate the ensembled prediction probability of the baseline and the SR models for each language. Then we compare the KL divergence between this ensemble distribution and \mvreg~with or without the consistency loss. In \autoref{fig:ensemble}, we plot this KL divergence difference between the \mvreg~without consistency loss and the full \mvreg for each language in NER. For most of the languages, the full  \mvreg~has lower KL divergence with the ensemble distribution, which indicates that the consistency loss trains the model to be closer to the ensemble of two inputs.

%

\subsection{\mvreg~also improves English}
\begin{figure}
    \centering
    \includegraphics[width=\columnwidth]{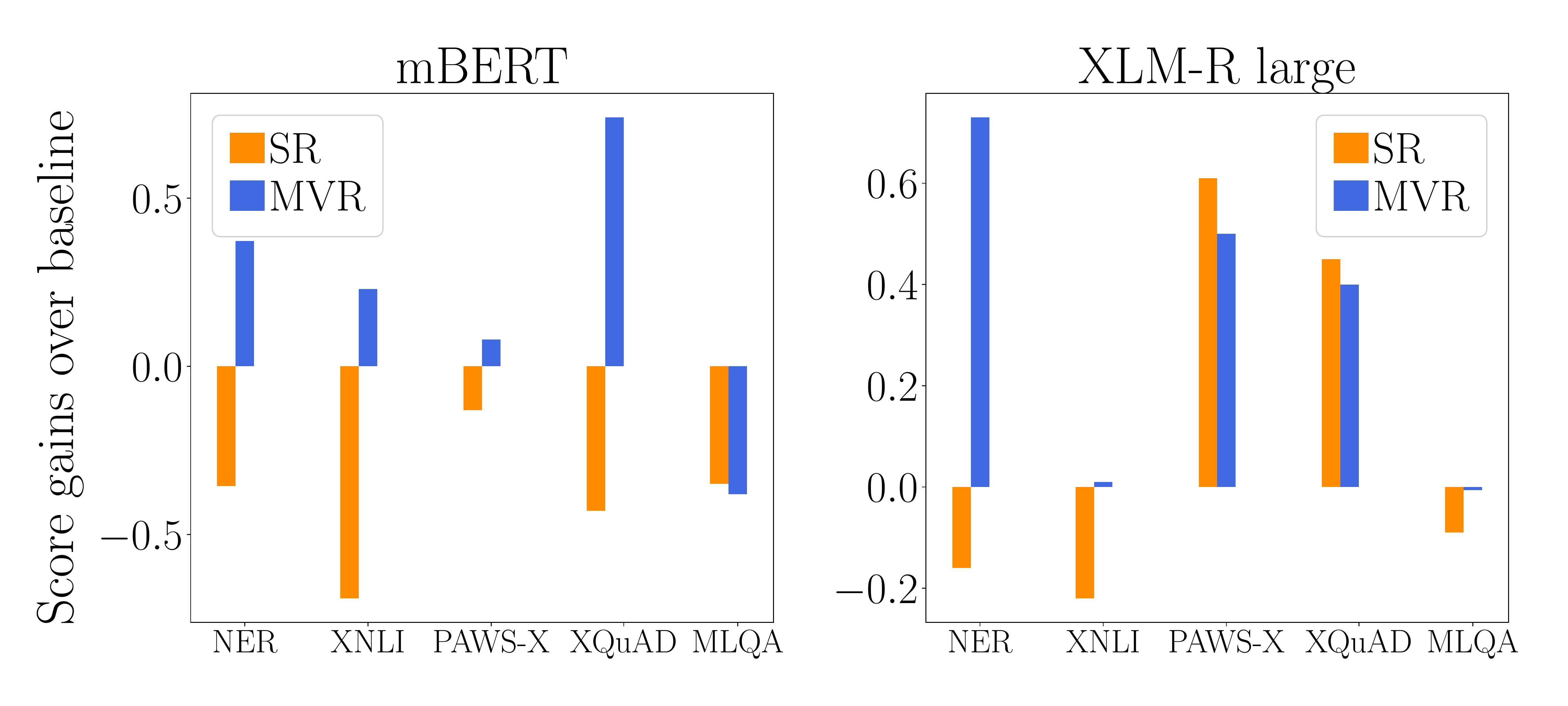}
    \caption{Gains of \mvreg~and SR for English. While \ps~harms the performance on English, \mvreg~generally improves it.}
    \label{fig:gain_for_en}
\end{figure}

Although \ps~improves the model performance averaged over all languages, surprisingly it can hurt the performance on English, the language we use for fine-tuning. \autoref{fig:gain_for_en} shows the improvement on English over the baseline for both \ps~and \mvreg, and notably English performance decreases for all tasks on mBERT. \mvreg, on the other hand, generally brings improvements for English across both mBERT and XLM-R large models. This is likely because \mvreg also utilizes English inputs with standard segmentation, the method used at pretraining time, which allows it to take full advantage of the information encoded during pretraining.  

\section{Related work \label{sec:relate}}
Several works propose to optimize subword-sensitive word encoding methods for pretrained language models. \citet{charbert} uses convolutional neural networks~\citep{kim-2014-convolutional} on characters to calculate word representations. \citet{ambert} propose to add phrases into the vocabulary for Chinese pretrained language models. However, they focus on improving the vocabulary of pretrained representations of a single language, and they require modification to the model pretraining stage. \citet{cluster_vocab_multilingual} propose to cluster related languages together and run subword vocabulary construction on each language cluster when constructing vocabularies for mBERT. Their method is also applied at the pretraining stage and could be combined with our method for potential additional improvements.  

Our method is also related to prior work that optimize word representations for NMT and language modeling. Character level embeddings have been utilized instead of subword segmentation for NMT~\cite{char_nmt_compression, fully_char_nmt,char_compose_nmt} and language modeling~\cite{char_lm,explore_limit_lm}. 
\citet{sde} propose a multilingual word embedding method for NMT that relies on character n-gram embedding and a latent semantic embedding shared between different languages. \citet{char_compose_nmt} show that character n-gram based embedding performs better than BPE for morphologically rich languages. \citet{he-etal-2020-dpe} propose to learn the optimal segmentation given a subword vocabulary for NMT.

Our method is inspired by semi-supervised learning methods that enforce model consistency on unlabeled data. Several self-training methods utilize unlabeled examples to minimize the distance between the model predictions based on the unlabeled example and a noised version of the same input~\cite{vat,adv_semi_text,xu-yang-2017-cross,cvt,uda}. \citet{xu-yang-2017-cross} use knowledge distillation on unlabeled data to adapt models to a new language. \citet{cvt} propose to mask out different parts of the unlabeled input and encourage the model to make consistent prediction given these different inputs. 
These methods all focus on semi-supervised learning, while our method regulates model consistency to mitigate the subword segmentation discrepancy between different languages. 
\section{Conclusion \label{sec:conclusion}}
We believe that the results in this paper convincingly demonstrate that standard deterministic subword segmentation is sub-optimal for multilingual pretrained representations. Even incorporating simple methods for subword regularization such as BPE-dropout at fine-tuning can improve the cross-lingual transfer of pretrained models, and our proposed Multi-view Subword Regularization method further shows consistent and strong improvements over a variety of tasks for models built upon different subword segmentation algorithms.
Going forward, we suggest that some variety of subword regularization, MVR or otherwise, should be a standard component of the fine-tuning of pre-trained representations that use subword segmentation.

\subsubsection*{Acknowledgments}
The first author XW is supported by the Apple PhD fellowship. This project is made possible by the computing resources from the Pittsburgh Supercomputing Center. The authors would like to thank Adhi Kuncoro for his comments regarding the draft of the paper.

\bibliography{anthology,custom}
\bibliographystyle{acl_natbib}

\clearpage
\appendix

\section{Appendix\label{sec:appendix}}

\subsection{Effect of hyperparameters\label{app:hyperparam}}
\begin{figure}
    \centering
    \includegraphics[width=0.7\columnwidth]{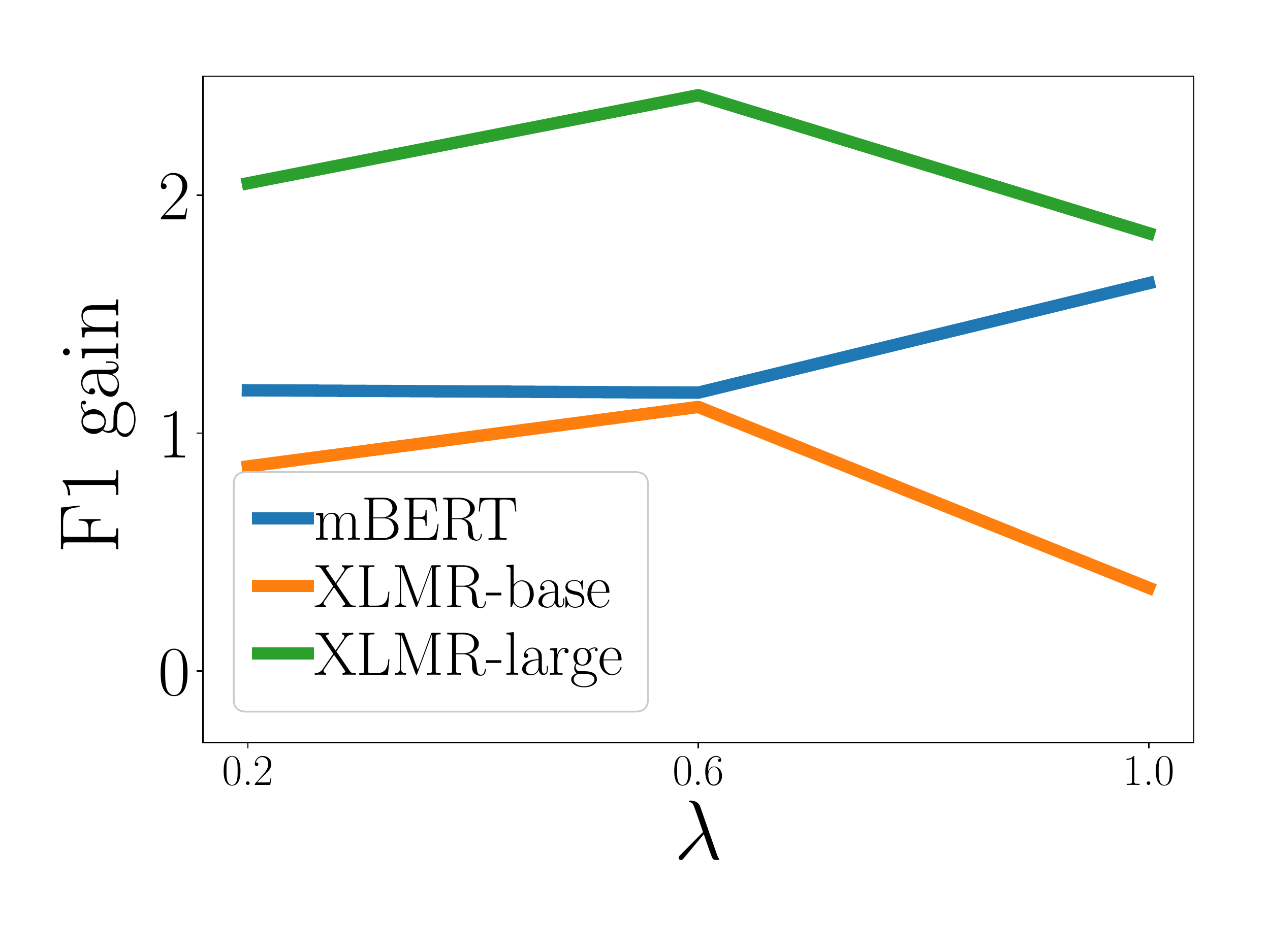}
    \caption{Average F1 gains over the baseline on NER task.}
    \label{fig:kl}
\end{figure}

In this section, we study the effect of two hyperparameters in \mvreg: the weight of consistency loss $\lambda$ and the temperature $\tau$ for flattening the prediction distribution. First, we analyze the effect of $\lambda$ on the NER tasks in \autoref{fig:kl}. We notice that mBERT performs better using a larger $\lambda$, while in general a value between 0.2 to 0.6 works reasonably well for all models. Note that we still use $\lambda=0.2$ for mBERT because we selected this value based on the performance on the English dev set.

\begin{figure}
    \centering
    \includegraphics[width=0.7\columnwidth]{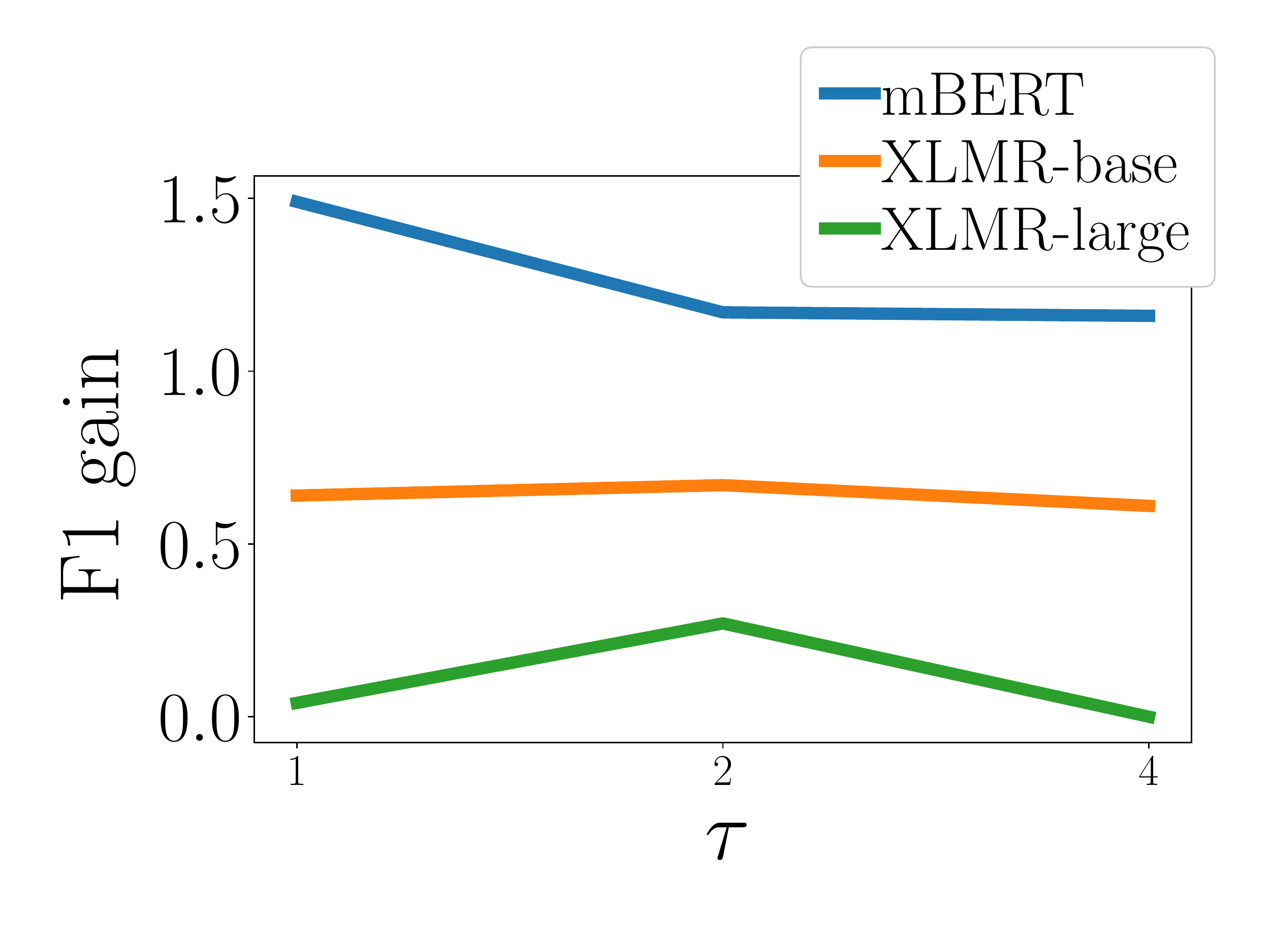}
    \caption{Average F1 gains over the baseline on QA tasks.}
    \label{fig:temp}
\end{figure}

We only tune the temperature $\tau$ for question answering tasks because it has a larger output space than other tasks. We plot the average F1 improvement on the QA tasks in \autoref{fig:temp}. 

Simplying using $\tau=1$ works well for the smaller mBERT and XLM-R base models. The best-performing XLM-R large model benefits from a larger $\tau$, or a flattened distribution, probably because its prediction distribution is relatively sharper or more confident than others. Generally the model is not particularly sensitive to the value of $\tau$.

Sequence tagging task requires truncating the inputs and targets to a predefined length. This could be a problem for calculating KL divergence when the deterministic and probabilistic segmented inputs have different number of tags been discarded. In our implementation of MVR, we simply calculate the KL divergence on the tags shared by the two inputs. Details of this implementation can be found in the code base. 


\subsection{Training details}
We select hyperparameters based on the validation performance of English on the NER task.
We fine-tune all models on the NVIDIA V100 GPU. Both SR and \mvreg~have the same number of model parameters as the baseline. Baseline experiments on all tasks except for the XNLI classification task generally finish within 5 hours on a single GPU. The baseline experiment on XNLI takes about 24 hours on 2 GPUs. SR takes about the same training time as the baseline, and \mvreg~takes about twice the amount of time.    

\subsection{Other analysis}
\begin{figure}
    \centering
    \includegraphics[width=0.9\columnwidth]{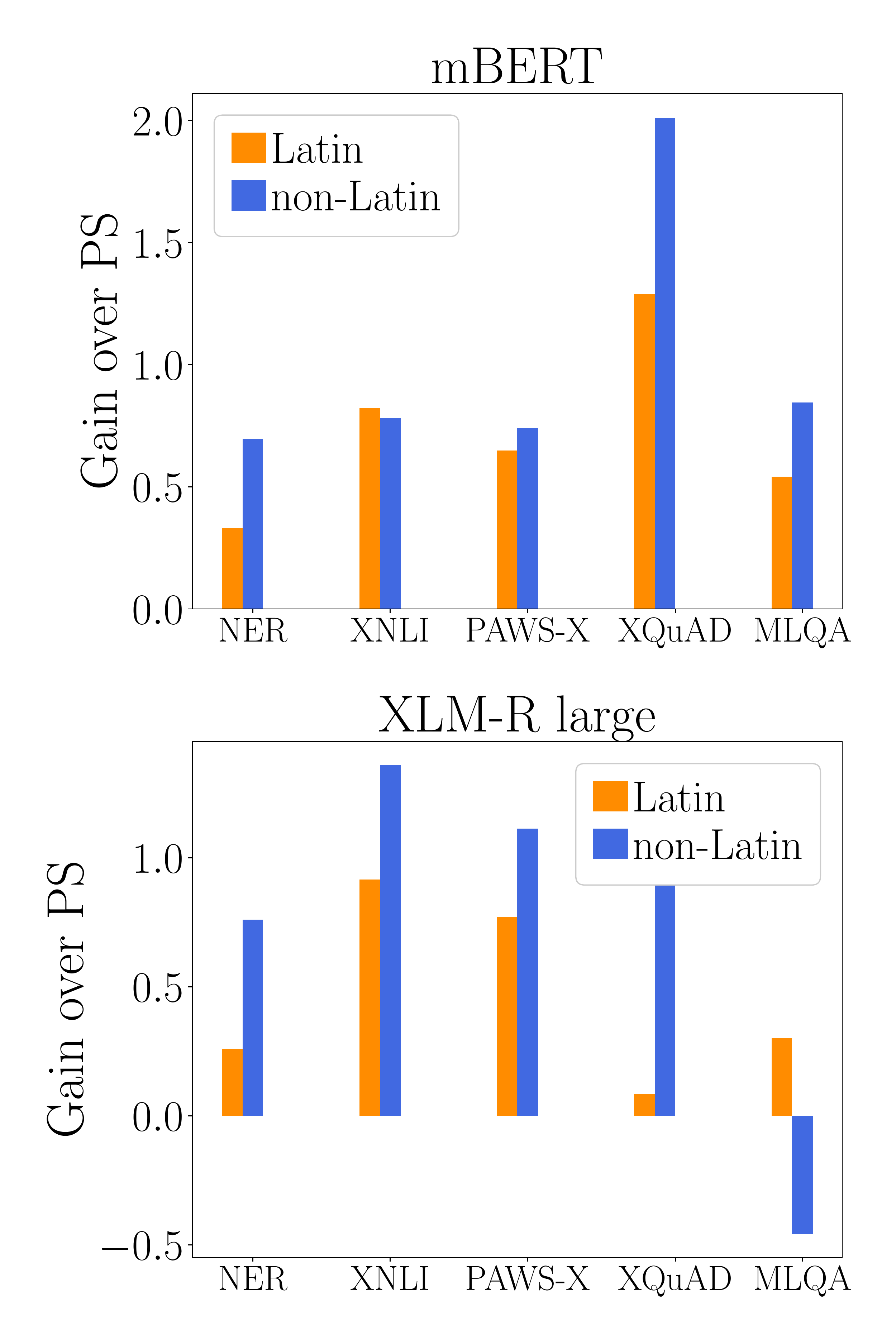}
    \caption{Gains of \mvreg~over SR for languages with Latin vs. non-Latin script.}
    \label{fig:gain_subword_for_script}
\end{figure}

\paragraph{\mvreg~improves more for languages with non-Latin script}
We further compare the gains of \mvreg~over SR on languages with Latin and non-Latin script. The plots can be found in \autoref{fig:gain_subword_for_script}. Overall \mvreg~leads to larger improvements over subword regularization for languages with non-Latin script for both mBERT and XLM-R large. By enforcing prediction consistency between different segmentation, \mvreg~is better than SR at making the model robust to languages with very different segmentation than the language used for finetuning.  

\end{document}